%% file: main.tex
\documentclass[runningheads]{llncs}

\usepackage{eccv}

\usepackage{eccvabbrv}

\input{preamble}

\usepackage{graphicx}
\usepackage{booktabs}
\usepackage{utfsym}
\newcommand{\NO}{ }
\newcommand{\YES}{\usym{1F5F8}}
\usepackage{array}
\newcommand{\PreserveBackslash}[1]{\let\temp=\\#1\let\\=\temp}
\newcolumntype{C}[1]{>{\PreserveBackslash\centering}p{#1}}
\newcolumntype{R}[1]{>{\PreserveBackslash\raggedleft}p{#1}}
\newcolumntype{L}[1]{>{\PreserveBackslash\raggedright}p{#1}}

\usepackage[accsupp]{axessibility}  

\usepackage[pagebackref,breaklinks,colorlinks]{hyperref}

\usepackage{orcidlink}

\begin{document}

\title{NToP: NeRF-Powered Large-scale Dataset Generation 
        for 2D and 3D Human Pose Estimation in Top-View Fisheye Images} 

\titlerunning{NToP}

\author{Jingrui Yu\inst{1} \and
Dipankar Nandi\inst{1} \and
Roman Seidel\inst{1} \and
Gangolf Hirtz\inst{1}}
\authorrunning{Yu et al.}

\institute{Technische Universität Chemnitz, 09126 Germany\\
\email{{\{jingrui.yu, dipankar.nandi, roman.seidel, g.hirtz\}}@etit.tu-chemnitz.de}\\
\url{https://www.tu-chemnitz.de/etit/dst/professur/index.php.en} }

\maketitle

\input{sections/abstract}

\input{sections/introduction}

\input{sections/relatedwork}
\input{sections/pipeline}
\input{sections/dataset}

\input{sections/experiments}
\input{sections/discussion}
\input{sections/conclusion}
\input{sections/supp}

%
%

\bibliographystyle{splncs04}
\typeout{}
\bibliography{literature}
\end{document}

%% file: preamble.tex
\usepackage[dvipsnames]{xcolor}
\newcommand{\red}[1]{{\color{red}#1}}


%% file: sections/abstract.tex
\begin{abstract}
Human pose estimation (HPE) in the top-view using fisheye cameras presents a promising and innovative application domain.
However, the availability of datasets capturing this viewpoint is extremely limited, especially those with high-quality 2D and 3D keypoint annotations.
Addressing this gap, we leverage the capabilities of Neural Radiance Fields (NeRF) technique to establish a comprehensive pipeline for generating human pose datasets from existing 2D and 3D datasets, specifically tailored for the top-view fisheye perspective.
Through this pipeline, we create a novel dataset NToP (\textbf{N}eRF-powered \textbf{To}p-view human \textbf{P}ose dataset for fisheye cameras) with over 570 thousand images, and conduct an extensive evaluation of its efficacy in enhancing neural networks for 2D and 3D top-view human pose estimation.
Extensive evaluations on existing top-view 2D and 3D HPE datasets as well as our new real-world top-view 2D HPE dataset OmniLab prove that our dataset is effective and exceeds previous datasets in this field of research.
The code and the trained models of NToP will be made available in the near future.
\keywords{NeRF \and human pose estimation \and fisheye \and top-view \and image synthesis}
\end{abstract}

%% file: sections/introduction.tex
\section{Introduction}

In recent years, indoor monitoring with an omnidirectional fisheye camera has become a growing application field of computer vision.
The large field-of-view of fisheye lenses enables the user to monitor an entire room with a single device by mounting it to the ceiling \cite{yu2023survey}.
Researchers have been successful in applying this top-view for the task of person detection by gathering large amounts of real-world data to train deep neural network based detectors \cite{yu2019omnipd,duan2020rapid,tezcan2022wepdtof}.
While person detection has reached state-of-the-art accuracies that are in line with implementations for normal perspective images, other applications of the top-view setup are hardly explored.
One of the most import application is human pose estimation (HPE).
Human keypoint information, in both 2D and 3D, is critical for accurately identifying emergency situations, such as falling of an elderly.
It also enables in-depth analysis of the actions and behaviors of the person \cite{degardin2021human,dhiman2019review,xin2023transformer}.
Until now, researchers have had to rely on rudimentary geometric information for fall detection \cite{delibasis2016falldetection,kottari2019real,siedel2020contactless,nguyen2021incorporation}, let alone solving the more complex challenge of high level behavior analysis.

The greatest barrier to solving HPE in top-view fisheye images is the lack of large-scale datasets.
Despite the abundance of HPE data in side-view \cite{h36m_pami,lin2014coco,andriluka14mpii}, real-world top-view HPE datasets are scarce.
Researchers are resorting to synthetic data generation \cite{yu2023poseomni,Garau2021panoptop,seidel2021omniflow} to mitigate this problem.
However, the domain gap between purely synthetic images or low-resolution images synthesized from real-world images and real-world images are limiting the performance.
The invention of NeRF \cite{mildenhall2020nerf} and the subsequent development of its human-centric variations \cite{weng2022humannerf,su2021anerf,peng2021neuralbody} open up new possibilities for high quality semi-synthetic HPE data generation.

\begin{figure}[t]
   \centering
   \includegraphics[width=\textwidth]{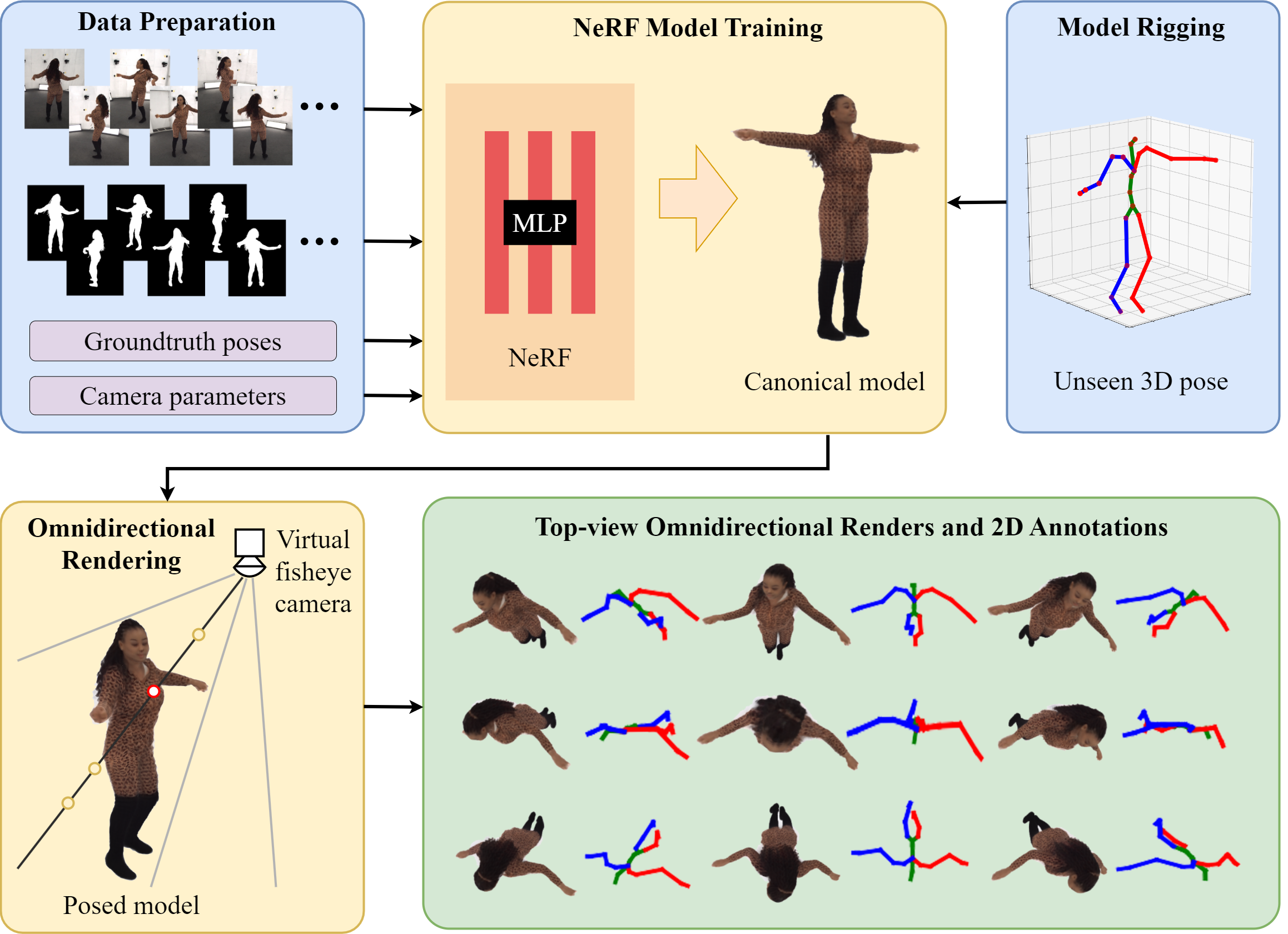}
   \caption{The NToP pipeline. We input images and their corresponding segmentation masks, groundtruth poses and camera parameters to HumanNeRF without the pose correction. After training, virtual fisheye cameras are positioned on top of the human model to render top-view images of a novel 3D pose. 2D groundtruth keypoint annotations are generated in post-processing.}
   \label{fig:pipeline}
\end{figure}

In this paper, we demonstrate that NeRF is suitable for the generation of highly realistic semi-synthetic omnidirectional top-view human pose images.
Our goal is to solve HPE in top-view fisheye images without re-inventing the wheels, leveraging the large bodies of available datasets and algorithms.
Our contributions are three-fold:
\begin{itemize}
    \item we present a general workflow (\cref{fig:pipeline}) for rendering omnidirectional top-view human images with virtual fisheye cameras and extracting corresponding groundtruth 2D and 3D keypoint annotations using a NeRF model,
    \item we introduce our large-scale HPE dataset, NToP, featuring over 570\,K high quality semi-synthetic human images in the top-view and groudtruth 2D and 3D keypoint annotations,
    \item we validate the effectiveness of our dataset by training and evaluating state-of-the-art single-view human pose estimators, namely ViTPose for 2D HPE \cite{xu2022vitpose} and HybrIK-Transformer for 3D HPE \cite{li2021hybrik,oreshkin2023hybrikT}, on existing datasets as well as our newly created dataset, OmniLab.
\end{itemize}



%% file: sections/relatedwork.tex
\section{Related Work}
\subsection{2D and 3D Human Pose Estimation with Deep Learning}
HPE, which refers to the task of locating the joints of one or multiple persons from images or videos, either in the image plane or in the 3D space, is one of the most important applications of computer vision, and thus widely researched.
Previous work in this area for normal perspective images in the side-view has been extensively covered by a number of surveys \cite{Liu2022posesurvey,Han2023transformersurvey,tian2022meshsurvey}.
Here we briefly take a look at different categories of the algorithms.
In the field of 2D HPE, there are two main approaches: the top-down approach and the bottom-up approach. 
The top-down approach takes a human instance as input, and directly outputs the possible joint locations.
It consists of a backbone CNN network for feature extraction and a pose estimation head that generates heatmaps for the joints.
The most widely used backbone networks include ResNet \cite{He2016resnet,xiao2018simple}, Hourglass \cite{newell2016hourglass} and HRNet \cite{Sun2019hrnet,cheng2020higerhrnet}.
The bottom-up approach takes the whole image and estimate the joints for multiple persons at the same time, which requires an extra algorithm for joint-person association \cite{pishchulin2016deepcut,cao2017realtime,newell2017associative}.
The powerful Transformer architecture has also progressed into HPE \cite{yuan2021hrformer,yang2021transpose,li2021tokenpose}, but ViTPose is the first to abandon the CNN backbone \cite{xu2022vitpose}.
It is superior in performance, training, scalability and transferability compared to CNNs.
3D pose estimation can be achieved using lifting methods \cite{martinez2017_3dbaseline}, but to definitively recover the 3D pose, temporal information \cite{motionbert2022}, multi-view \cite{iskakov2019learnable} or latent encoding of the human body \cite{Shetty2023PLIKS} is necessary.
Researchers also combine the above methods for better performance \cite{Liu2022posesurvey}.

\subsection{Top-View Human Pose Estimation Algorithms and Datasets}

Haque \etal \cite{haque2016vpinvariant3dhpe} is one of the earliest implementations of deep learning for view-invariant pose estimation.
They utilize a CNN for body part detection and an LSTM for pose refinement in depth images, which they presented in their ITOP dataset.
Garau \etal use capsule autoencoders to achieve view-invariant pose estimation in RGB images and depth maps \cite{garau2021deca}.
They created the PanopTOP31K dataset by image synthesis \cite{Garau2021panoptop} from 3D point clouds that are available in the Panoptic dataset \cite{Joo2015panoptic}.
It features corresponding RGB image and depth map pairs from the top-view and the side-view.
Denecke and Jauch \cite{denecke2021verification} leverage the prior knowledge of human anatomies to analyze the quality of top-view HPE by a neural network.
The analysis results can be used to improve the neural network.
Heindl \etal \cite{heindl2019large} triangulate keypoints that are detected by OpenPose \cite{cao2017realtime} in rectified fisheye images with the help of camera parameters.
Yu \etal \cite{yu2023poseomni} created a synthetic dataset THEODORE+ and finetuned CNN-based pose estimators for direct keypoint detection in top-view fisheye images.
They evaluate their method on a real-world dataset PoseFES to demonstrate the improvement in critical cases where algorithms for side-view fail.

\begin{table}[t]
    \centering
    \small
    \caption{Datasets for top-view human pose estimation.}
    \label{tab:topviewdatasets}
    \scriptsize
    \begin{tabular}{lcrccccccc}
        \toprule
        Dataset & type & frames & RGB & SegMask & SideView  & 2DGT & 3DGT & CamParams & Depth \\
        \midrule
        ITOP & real & 50\,K & \NO & \NO & \NO & \NO & \YES & \YES & \YES \\
        PanopTOP31K & semi-syn & 31\,K & \YES & \NO & \YES & \YES & \YES & \YES & \YES \\
        THEODORE+ & synthetic & 50\,K & \YES & \YES & \NO & \YES & \YES & \NO & \NO \\
        PoseFES & real & 0.7\,K & \YES & \NO & \NO & \YES & \NO & \NO & \NO \\
        WEPDTOF-pose & real & 6.7\,K & \YES & \NO & \NO & \YES & \NO & \NO & \NO \\
        \midrule
        \textbf{NToP} (ours) & semi-syn & 570\,K & \YES & \YES & \YES & \YES & \YES & \YES & \NO \\
        \textbf{OmniLab} (ours) & real & 4.8\,K & \YES & \NO & \NO & \YES & \NO & \NO & \NO \\
        \bottomrule
    \end{tabular}
\end{table}

It is noticeable that most publications mention creating one or more datasets for training or evaluation.
We summarize the top-view datasets in \cref{tab:topviewdatasets}.
Besides the publications mentioned above, the CMU Panoptic dataset \cite{Joo2015panoptic} contains over-head recordings, though not directly from the top.
Similarly, the MPI-INF-3DHP dataset \cite{Mehta3dhp2017} is recorded with multiple views, two of which are from over-head positions.
In addition, recent motion capture datasets such as Genebody \cite{cheng2022genebody} and DNA-Rendering \cite{cheng2023dnarendering} also include images captured from a higher position than the human head, yet the cameras are not downward facing.

\subsection{NeRF and Human-Centric NeRF Variants}

Neural radiance field (NeRF) is a new way for high quality novel view synthesis from sparse views and has gained enormous attention since its invention \cite{mildenhall2020nerf}.
It models a continuous scene in the 3D space as a function \(F_{\Theta}: (\mathbf{x},\mathbf{d}) \rightarrow (\mathbf{c},\sigma)\), where \(\mathbf{x}\) is a position in space, \(\mathbf{d}\) is a looking direction of a light ray, \(\mathbf{c}\) is the RGB color and \(\sigma\) is the volumetric density.
The weights \(\Theta\) of a multilayer perceptron (MLP) is trained to learn \(F\).
In rendering, the color \(C\) of a pixel can be calculated by the integral of \(\mathbf{c}\) and \(\sigma\) along the light ray \(\mathbf{r}(t)=\mathbf{o}+t\mathbf{d}\) from the near bound \(t_n\) to the far bound \(t_f\):
\begin{equation}
    C(\mathbf{r}) = \int_{t_n}^{t_f} T(t)\sigma\bigl (\mathbf{r}(t)\bigr )\mathbf{c}\bigl (\mathbf{r}(t),\mathbf{d}\bigr )dt
    \label{eq:nerfrender}
\end{equation}
\begin{equation}
    \text{where}\quad T(t) = \exp \Bigl (-\int_{t_n}^{t}\sigma\bigl (\mathbf{r}(s)\bigr )\Bigr )ds
    \label{eq:nerftt}
\end{equation}
\(T(t)\) is the accumulated transmittance along the ray from \(t_n\) to \(t\).
NeRF essentially encodes a 3D scene including the lighting into an MLP.
When presented with sufficient number of views that surround the scene, NeRF can learn the whole scene and enables rendering from any direction in the 3D world.

Apparently, NeRF is restricted to static scenes and can not be applied to moving humans in its vanilla implementation \cite{peng2021neuralbody,2021narf,su2021anerf}.
NARF \cite{2021narf} extends NeRF to articulated objects, particularly to the human body parameterized by a 3D skeleton.
The density $\sigma$ is omitted because the human body is inherently not transparent.
A-NeRF \cite{su2021anerf} is a similar implementation to NARF.
NeuralBody \cite{peng2021neuralbody} uses SMPL instead of the simple skeleton as latent encoding of the human body.
SMPL \cite{SMPL2015} is a learned human mesh model that simulate both the rigid transformation of body parts and the soft deformation of the human tissue realistically, given the skeleton pose.
SMPL has become the default go-to for many tasks related to 3D human reconstruction.
H-NeRF \cite{xu2021hnerf} uses an MLP-based statistical human model imGHUM \cite{xu2020ghum,alldieck2021imghum} instead of SMPL for the latent encoding.
HumanNeRF \cite{weng2022humannerf} adds non-rigid deformation into consideration and achieves realistic rendering using a single input video.
NeuMan \cite{jiang2022neuman} concurrently train two NeRFs, one for the human and the other for the scene, from a single video, and enables rendering the human with new poses in the original scene.
TAVA \cite{Li2022tava} adds a neural blend skinning function to let the MLP learn the deformation directly, thus eliminating the need for a human model.
Recent advances focus on reducing temporal cost \cite{Jiang2023instantavatar,Geng2023CVPR,zheng2023avatarrex,chen2023fastsnarf}, reducing supervision \cite{Guo2023vid2avatar,Siarohin2023CVPR,Chen2023CVPR,hu2023sherf} and enabling multiple usage of a single trained model \cite{wang2023clothednerf,mu2023actorsnerf}.

%% file: sections/pipeline.tex
\section{NToP Data Generation Pipeline}
The NToP data generation pipeline consists of three major steps:
NeRF model training, omnidirectional rendering with virtual fisheye cameras and the generation of groundtruth 2D annotations, see \cref{fig:pipeline}.
In the following subsections, we explain the steps in detail.

\subsection{NeRF Model Training}
The first overall step in the pipeline is NeRF model training.
We tackle the various aspects of this task as follows:

\textbf{Origin Dataset Selection.}
The origin dataset for training the NeRF model should meet at least three criteria: 1) the camera intrinsic and extrinsic parameters are known, 2) all sides of the person are captured during the recording, and 3) there are groundtruth 3D human pose annotations.
In addition, it is beneficial if the capture system includes one or more cameras that are positioned higher than the subject's head.
For example, the Human3.6M dataset fulfills these criteria, because it contains videos from four cameras and their corresponding parameters.
The human pose annotation is acquired by a motion capture system, and given as keypoint annotations in 3D.

\textbf{Data Preprocessing.}
We take Human3.6M as an example to explain the necessary steps.
First, the videos are extracted into image sequences.
Second, the person is extracted from the background using background subtraction or a segmentation mask generator, such as the segment anything model (SAM) \cite{kirillov2023segany}.
In some cases, the segmentation mask must be corrected manually.
Third, the 3D groundtruth keypoints are mapped to an SMPL model using MoSh++ \cite{Loper2014mosh,2019amass}.
Specifically, the SMPL parameters are ten shape parameters \(\mathbf{\beta}\) and the poses as a \(24\times3\) matrix.
More recent motion capture datasets such as those in AMASS \cite{2019amass} use the SMPL-X \cite{SMPL-X:2019} model as the unified annotation format, thus this step can be omitted.
In this step, the skeleton is normalized, and the root joint \emph{pelvis} is aligned to the origin of the coordinate system \((0,0,0)\).
The preprocessed data should contain 200 to 800 frames for \emph{each actor}, the corresponding camera intrinsic and extrinsic parameters, the person segmentation masks, as well as the SMPL models for \emph{each frame}.
Notably, the training set does not need to cover the render range, and the images are not necessarily from the same camera.
For example, for actor S1 in Human3.6M, only the images from one action is used for training the NeRF model, since the clothing does not change across all actions.
Images from all four cameras are used to generate the training set, so that the trained model contains all sides of the actor.

\textbf{Model Training.}
We select the framework proposed by Weng et al. \cite{weng2022humannerf} as the foundation for training our human models, leveraging its impressive rendering quality.
Despite the original training methodology relying on single-view videos, we utilize known camera parameters for each frame, facilitating training with a multi-view dataset.
Notably, we opt out of using the pose correction module and instead provide ground truth annotations.
One NeRF model is trained for each actor using the preprocessed data.
A typical training session trains for 100k steps with a learning rate of 0.0005, batch size of 1, number of patches of 36, using the adam optimizer.
These parameters are suitable for a workstation with two Nvidia Titan RTX graphics cards when training on images of the resolution \(1000\times 1000\) px.


\subsection{Fisheye Camera Model and Omnidirectional Rendering}
\label{subsec:cammodel}

\begin{figure}[t]
    \centering
    \subcaptionbox{\label{subfig:fisheyemodel}}[.5\linewidth]{\includegraphics[width=\linewidth]{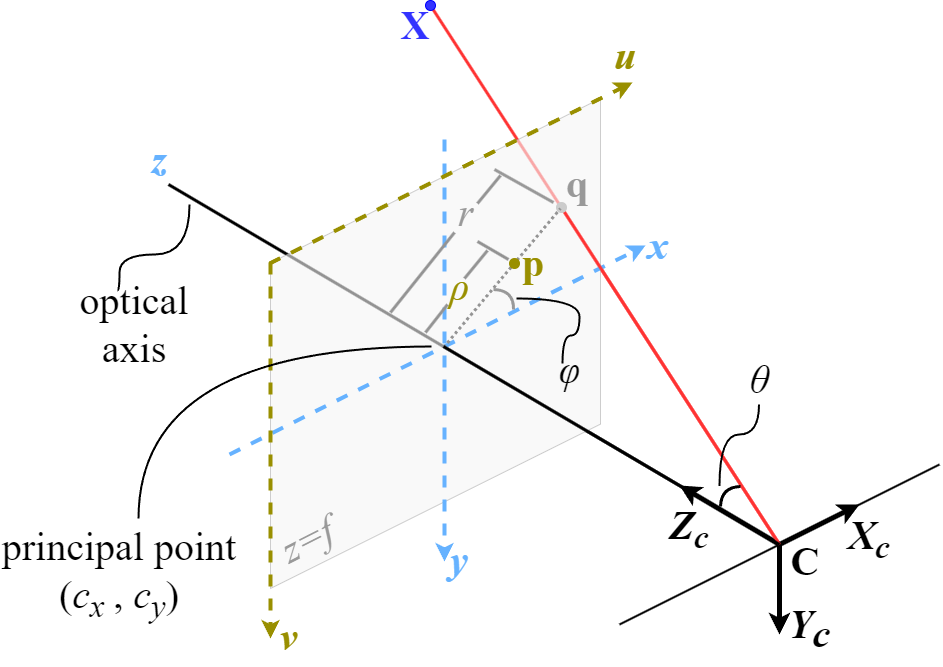}}
    \hspace{.5cm}
    \subcaptionbox{\label{subfig:rays}}[.36\linewidth]{\includegraphics[width=\linewidth]{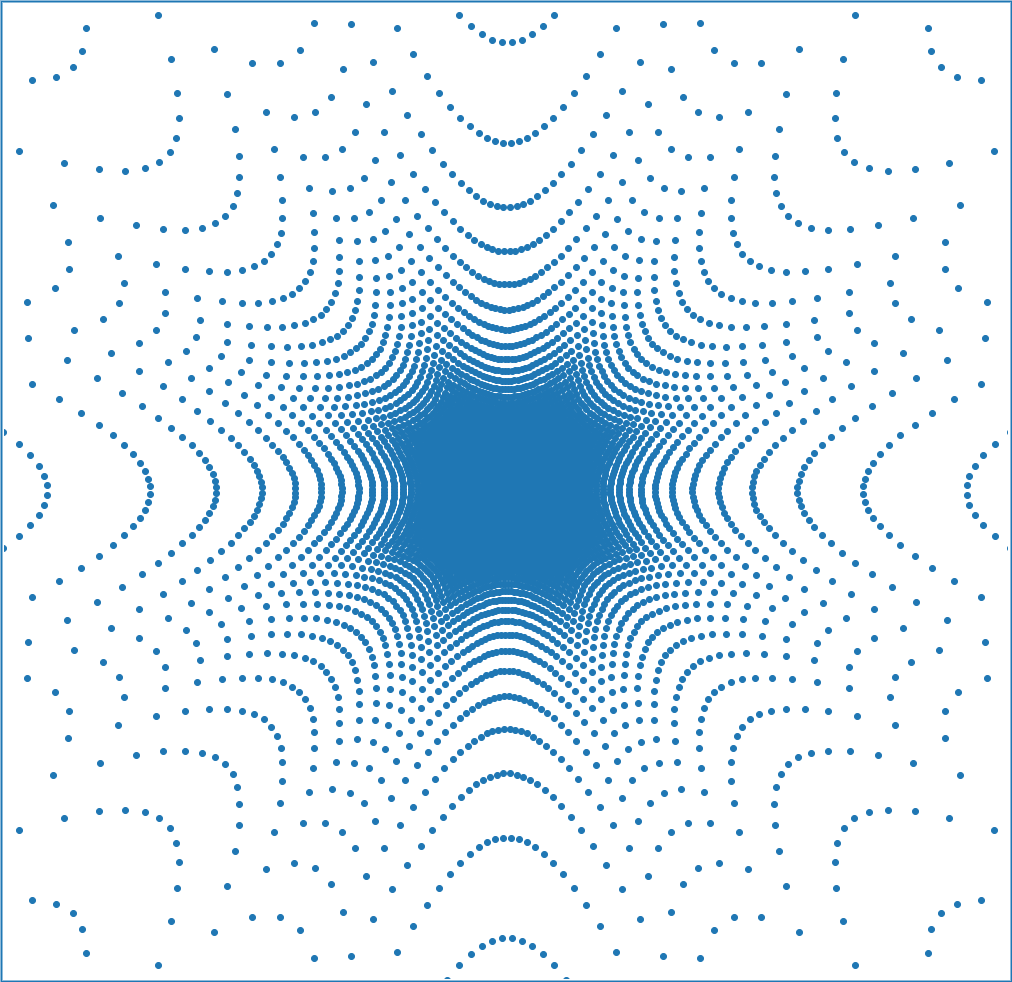}}
    \caption{(a) The equidistant projection model. \(C\) is the camera center, \((\boldsymbol{X}_\mathbf{C},\boldsymbol{Y}_\mathbf{C},\boldsymbol{Z}_\mathbf{C})\) is the CCS. (b) Distribution of the ray cross points \(\mathbf{q}\)  on the image plane for a demonstrative \(50\times 50\) pixel omnidirectional render. Extreme far points are not plotted. }
    \label{fig:fisheyemodel}
\end{figure}

\textbf{The Forward Projection.}
In \cref{fig:fisheyemodel}, let \(\mathbf{X}\) be a point in the 3D world, and its position in the world coordinate system (WCS) is noted as \(\mathbf{X}_\mathrm{w}\).
The extrinsic parameters of a camera is defined by its rotation matrix \(\mathbf{R}\) and translation matrix \(\mathbf{T}\) in WCS.
The position of \(\mathbf{X}_\mathrm{w}\) can be converted to the camera coordinate system (CCS) by \cref{eq:w2cproj}.
\begin{equation}
    \mathbf{X}_\mathrm{c} = \mathbf{R}\mathbf{X}_\mathrm{w} + \mathbf{T}
    \label{eq:w2cproj}
\end{equation}
The light ray that goes through \(\mathbf{X}\) and the camera center \(\mathbf{C}\) crosses the mirrored image plane at \(\mathbf{q}\) with the distance \(r\) to the principal point, and its elevation \wrt the optical axis of the camera is \(\theta\).
The azimuth \(\varphi\) is the angle between the \(x\)-axis and \(r\).
The resulting image pixel \(\mathbf{p}=(u,v)\) has the same azimuth \(\varphi\) as \(\mathbf{q}\), yet its distance to the principal point \(\rho\) is determined by the projection model.
The distortionless ideal fisheye camera for our rendering uses the equidistant projection, where \(\rho=f\theta\). 
We can then calculate the pixel coordinates of \(\mathbf{p}\) with \cref{eq:pixelcoords}, where \((c_x,c_y)\) are the coordinates of the principal point in the image coordinate system \((\boldsymbol{u,v})\).
\begin{equation}
    u = \rho\cos\varphi + c_x \qquad v = \rho\sin\varphi + c_y
    \label{eq:pixelcoords}
\end{equation}

\textbf{The Backward Ray Finding Process.}
The rendering pass in \cref{eq:nerfrender,eq:nerftt} requires that the ray for each pixel is known, which is the reverse calculation of the forward projection.
Given a pixel \(\mathbf{p}=(u,v)\), we need to calculate its corresponding cross point \(\mathbf{q}=(q_x,q_y,q_z)\) to find the ray.
We first calculate the distance between the image pixel and the principal point \(\rho\):
\begin{equation}
    \rho = \sqrt{(u-c_x)^2 + (v-c_y)^2}
\end{equation}
then get 
\begin{equation}
    r = f\tan\theta = f\tan\dfrac{\rho}{f}
\end{equation}
Using similar triangles, we calculate the coordinates of \(\mathbf{q}\) in CCS:
\begin{equation}
    q_x=\dfrac{r}{\rho}(u-c_x) \qquad q_y=\dfrac{r}{\rho}(v-c_y) \qquad q_z=f
\end{equation}
and convert them to WCS:
\begin{equation}
    \mathbf{q}_\mathrm{w} = (\mathbf{q} - \mathbf{T})\mathbf{R}
\end{equation}

The ray \(\mathbf{r}(t)=\mathbf{o}+t\mathbf{d}\) consists of its origin \(\mathbf{o}\), which is the camera center \(\mathbf{C}\), and the directional vector \(\mathbf{d}\).
Thus, we can finally calculate the ray with:
\begin{align}
    \mathbf{o} &= \mathbf{C} = -\mathbf{R}^\top \mathbf{T}\\
    \mathbf{d} &= \mathbf{q}_\mathrm{w} - \mathbf{o}
\end{align}
The distribution of the ray crossing points \(\mathbf{q}\) on the image plane is inherently non-linear. An example \(50\times 50\) render image is illustrated in \cref{subfig:rays}.

\textbf{Set Up Virtual Cameras for Rendering.}
With this, we only have to define the extrinsic parameters of the camera for rendering in this way:
\begin{enumerate}
    \item Set \(\mathbf{R}\) so that the camera faces downwards with its optical axis parallel to the vertical axis of WCS.
    \item The camera is given the initial translation matrix \(\mathbf{T}=(x_p, y_p, h)^\top\), in which \(x_p\) and \(y_p\) are the \(x\)- and \(y\)-coordinates of the root joint of SMPL, namely the pelvis, and \(h\) is the height of the camera.
    \item The camera is moved to the surround positions at E, NE, N, NW, W, SW, S and SE in a circle with radius \(R\) by setting its translation matrix to \(\mathbf{T}' = \mathbf{T} + \mathbf{T}_\mathrm{add}\), in which \(\mathbf{T}_\mathrm{add}=(R\cos{(n\pi/4)},\; R\sin{(n\pi/4),\; 0})^\top\), for \(n\in [0, 1, \ldots, 7]\).
    \item If more than one set of synthetic images are wanted for each origin frame, change \(h\) and \(R\) and repeat steps 2 and 3.
\end{enumerate}

\subsection{Groundtruth Keypoint Annotation}
By default, the 24 SMPL joints are used as groundtruth 3D keypoint annotations.
However, it is possible to convert them to other formats by inferring from the SMPL vertices.
Using the forward projection process in \cref{subsec:cammodel}, the coordinates of the 3D keypoints can be projected into the image coordinate system to form the groundtruth 2D keypoint annotations for each image.

%% file: sections/dataset.tex
\section{NToP Dataset and OmniLab Dataset}

\subsection{Origin Datasets and Rendering Parameters of NToP}
\textbf{Human3.6M} \cite{h36m}. We choose Human3.6M since it is the most used benchmark dataset in the field.
For the actors S1, S5, S6, S7, S8, S9, S11, which are available for download with groundtruth annotations, we use the image sequences from the action ``Posing'' for model training.
The original image sequences from all four cameras in Human3.6M are downsampled by the factor five and concatenated as the training dataset.
All 15 actions are used for rendering, but only one sub-action is taken.
Same as training set, the original image sequences are downsampled by the factor five.
Each input frame is rendered with \(h=1.2\) and \(R=1.0\) at the original resolution of \(1000\times1002\).
The subset generated from it is named \emph{ntopH36M} in the following sections.

\textbf{GeneBody} is a high resolution motion capture dataset that features actors of different genders, ages, body types and varied clothing \cite{cheng2022genebody}.
The original image sequences are already unified to 150 frames for each actor.
We use the images from cameras 01, 10, 19, 28, 37, 46 to form the training set for each actor, and trained 27 actors for rendering.
Two rendering passes are done for \(h=1.2\), \(R=1.0\) and \(h=1.0\), \(R=0.5\).
The input image resolution is \(2448\times2048\), but we rendered at the half resolution of \(1224\times1024\) to reduce rendering time and to keep the subjects roughly the same size as the Human3.6M renders.
The subset generated from it is named \emph{ntopGB} in the following sections.

\textbf{ZJU-MoCap} is a motion capture dataset with 9 sequences \cite{peng2021neuralbody}.
Training data is readily available as this dataset is used by multiple human body NeRF algorithms.
We render at \(1024\times1024\) for \(h=1.2\), \(R=1.0\) and \(h=1.0\), \(R=0.5\) for all sequences, which has been downsampled by factor five.
The subset generated from it is named \emph{ntopZJU} in the following sections.

During rendering, the subject mask is saved concurrently with the RGB image.
An annotation file that contains the SMPL parameters, the 3D keypoints and the parameters of the virtual cameras is written for each 9-frame render set.
The 2D groundtruth keypoints are generated in a post-processing step.

\subsection{OmniLab Dataset}
\begin{figure}[h]
    \centering
    \subcaptionbox{\label{subfig:ol1}}[.24\linewidth]{\includegraphics[width=\linewidth]{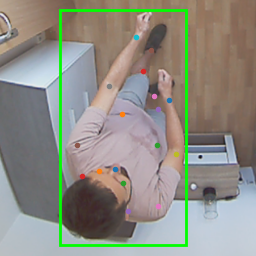}}
    \subcaptionbox{\label{subfig:ol2}}[.24\linewidth]{\includegraphics[width=\linewidth]{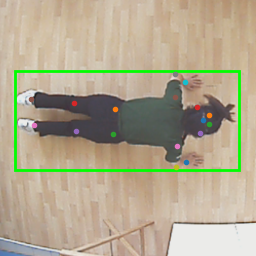}}
    \subcaptionbox{\label{subfig:ol3}}[.24\linewidth]{\includegraphics[width=\linewidth]{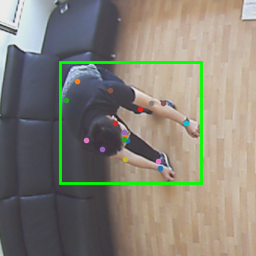}}
    \subcaptionbox{\label{subfig:ol4}}[.24\linewidth]{\includegraphics[width=\linewidth]{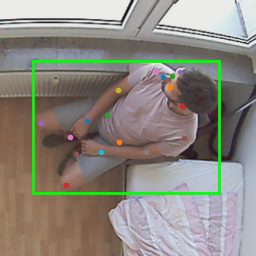}}
    \caption{Examples from OmniLab dataset. Actions: (a) brooming, (b) getting up from ground, (c) pulling object, and (d) sitting down and standing up.}
    \label{fig:omnilab}
\end{figure}

In order to evaluate the effectiveness of NToP in real-world scenarios, we collect a new dataset OmniLab with a top-view omnidirectional camera, mounted on the ceiling of two different rooms at \(2.5\)\,m height.
Five actors (3 males, 2 females) perform 15 actions from CMU-MoCap database in two rooms with varying clothes.
The recorded action length is 2.5\,s, which results in 60 images for each scene at a frame rate of 24\,FPS.
The position of the camera is fixed and the resolution of the images is \(1200\times1200\) pixels.
A total of 4800 frames are collected.
All annotations of 17 keypoints conforming to COCO conventions are estimated through a keypoint detector \cite{yu2023poseomni} and subsequently refined by four different humans in two loops to ensure high annotation quality.
\cref{fig:omnilab} shows a few examples from OmniLab with person bounding boxes and keypoint annotations.

\subsection{Dataset statistics and comparison}

\begin{figure}
    \centering
    \includegraphics[width=\textwidth]{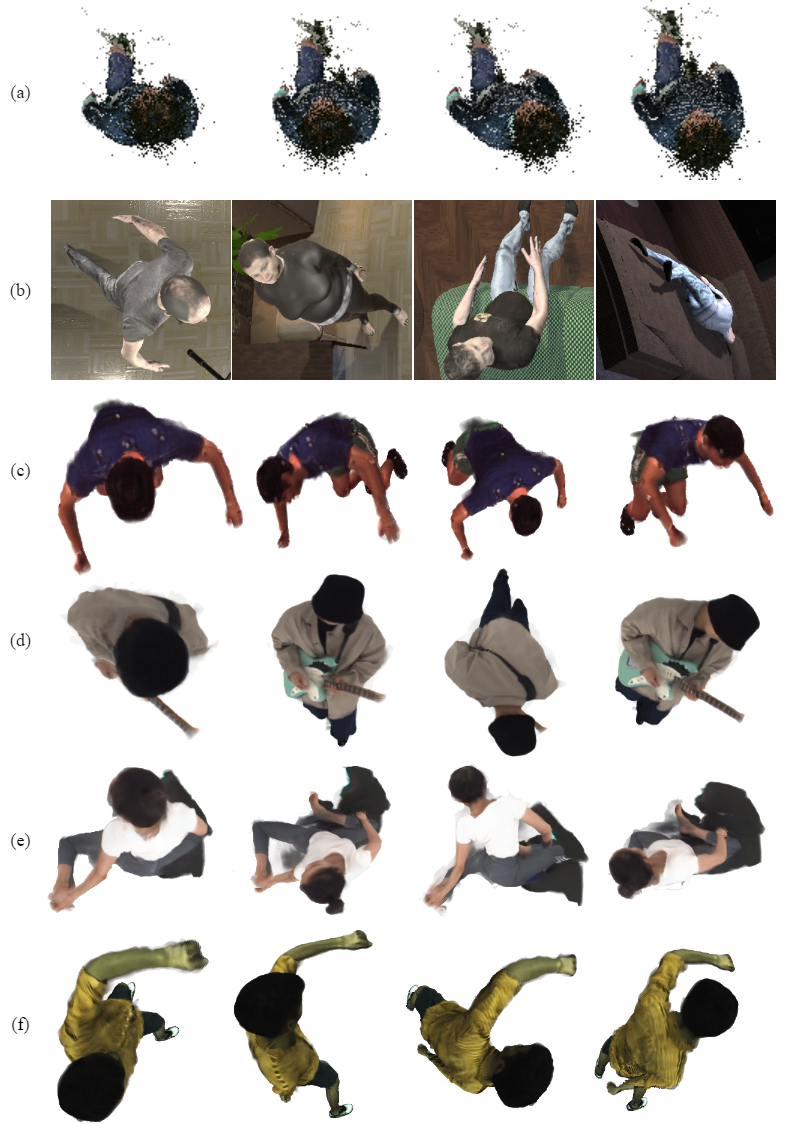}
    \caption{Dataset examples: (a) PanopTOP31K, (b) THEODORE+, (c) ntopH36M, (d,e) ntopGB, (f) ntopZJU. The subjects are resized to roughly the same size to showcase the difference in render quality.}
    \label{fig:rendereg}
\end{figure}

NToP is split into subsets based on the origin dataset, and each subset is split into train and validation by choosing different actors.
All the train and val subsets are concatenated into a total train and val set.
The statistics of NToP and OmniLab are summarized into \cref{tab:statistics} and compared to THEODORE+, PanopTOP31K and PoseFES.
Some qualitative results are provided in \cref{fig:rendereg}.

\begin{table}[t]
    \centering
    \small
    \caption{Statistics of datasets. The average subject size is measured by the subject bounding box. Actors refer to the unique persons in the origin dataset, and subjects refer to the rendered human instances.}
    \label{tab:statistics}
    \scriptsize
    \begin{tabular}{llrc}
        \toprule
        Dataset & Actors & Subjects & Avg. subject size \\
        \midrule
        THEODORE+ & - & \(\sim\)160\,K & 197\(\times\)213 \\
        PanopTOP31K & 24 & 31\,K & 102\(\times\)63 \\
        PoseFES & 7 & 2.9\,K & 219\(\times\)240 \\
        \midrule
        ntopH36M-train & 5 (2M, 3F) & 347\,K & 188\(\times\)187 \\
        ntopH36M-val & 2 (2M) & 123\,K & 199\(\times\)198 \\
        ntopGB-train & 19 (13M, 6F) & 51.3\,K & 270\(\times\)250 \\
        ntopGB-val & 8 (5M, 3F) & 21.6\,K & 263\(\times\)253 \\
        ntopZJU-train & 7 (7M) & 22.9\,K & 262\(\times\)244 \\
        ntopZJU-val & 2 (2M) & 4.1\,K & 262\(\times\)256 \\
        \midrule
        NToP-train & 31 (22M, 9F) & 421\,K & 202\(\times\)198 \\
        NToP-val & 12 (9M, 3F) & 149\,K & 210\(\times\)208 \\
        NToP total & 43 (31M, 12F) & 570\,K & 204\(\times\)201 \\
        \midrule
        OmniLab & 5 (3M, 2F) & 4.8\,K & 209\(\times\)191 \\
        \bottomrule
    \end{tabular}
\end{table}

We compare NToP with THEODORE+ and PanopTOP31K because they are the only two datasets created specifically for the top-view HPE task that features 2D and 3D groundtruth annotations and RGB images.
NToP is superior both quantitatively and qualitatively.
It has 18 times more subjects than PanopTOP31K and 3.5 times more than THEODORE+.
In PanopTOP31K, each subject is rendered with four camera positions, however, the variation of perspective is very small, which is visible from \cref{fig:rendereg}\textcolor{red}{a}, while NToP provides 9 or 18 varied perspectives for each subject.
THEODORE+ on the other hand has no control over the perspectives due to its randomized rendering procedure \cite{yu2023poseomni}.
Another advantage of NToP is the bigger subject size, see \cref{tab:statistics}.
In terms of quality, our new dataset is significantly more realistic than both other datasets and exhibits far less noise compared to PanopTOP31K.

%% file: sections/experiments.tex
\section{Dataset Validation}
\subsection{2D Pose Estimation with ViTPose}
\label{subsec:2dvitpose}

For the 2D HPE task, ViTPose sets a new Pareto front with the ViTPose-B, -L and -G models \cite{xu2022vitpose}.
The ViTPose-B model achieves the average precision (AP) of 77.1\,\% and the average recall (AR) of 82.2\,\% on MS COCO keypoints 2017 validation dataset at the resolution of \(256\times192\).
We use this as the baseline and finetune it with NToP-train and its subsets, as well as THEODORE+, alongside MS COCO 2017 train.
During training, NToP is augmented with random images from MIT Indoor Scenes database \cite{mitindoor} as background.
The finetuned models are evaluated on the val sets of NToP, PoseFES and OmniLab, and the results are summarized in \cref{tab:vitval}.
From the test results, we make the following observations:
1) The baseline model suffers significant performance drop on the val sets of NToP, but finetuning on \emph{any} of the NToP train subsets increases the performance on \emph{all} val subsets.
2) Finetuning the baseline model on THEODORE+ improves its performance on NToP val sets marginally.
3) There is a domain gap between NToP\,/\,THEODORE+ and real-world images, as described in \cite{yu2023poseomni}. Adding COCO to the training compensates this gap and the best performance is achieved by NToP-train.
In summary, our dataset shows clear advantage over the pure synthetic THEODORE+ dataset for 2D.

\begin{table}[t]
    \centering
    \scriptsize
    \caption{Evaluation results of 2D keypoints with ViTPose-B model on NToP-val and OmniLab. AP and AR are calculated following the MS COCO evaluation pipeline and presented in \%. }
    \label{tab:vitval}
    \begin{tabular}{lC{.8cm}C{.8cm}C{.8cm}C{.8cm}C{.8cm}C{.8cm}}
    \cmidrule[1pt]{2-7}
      & \multicolumn{2}{c}{NToP-val} & \multicolumn{2}{c}{OmniLab} & \multicolumn{2}{c}{PoseFES} \\ 
     \cmidrule{2-7}
    Training set   & AP   & AR     & AP   & AR  & AP   & AR\\
    \midrule
    baseline ViTPose-B & 46.36 & 49.82 & 76.62 & 78.53 & 54.20 & 57.70\\
    \midrule
    THEODORE+ & 54.44 & 57.74 & 69.14 & 72.56 & 46.90 & 51.23\\
    THEODORE+ \& COCO & 60.27 & 63.55 & 76.52 & 79.02 & 55.58 & 58.82\\
    \midrule
    ntopH36M-train & 73.51 & 75.93 & 61.32 & 64.95 & 36.42 & 40.54\\
    ntopGB-train & 63.60 & 67.60 & 74.95 & 77.71 & 53.56 & 57.44 \\
    ntopZJU-train & 57.15 & 61.87 & 69.58 & 73.59 & 52.38 & 46.48\\
    NToP-train & \textbf{79.71} & \textbf{82.07} & 70.19 & 73.49 & 46.84 & 51.34\\
    NToP-train \& COCO & 79.54 & 81.95 & \textbf{78.25} & \textbf{80.51} & \textbf{56.47} & \textbf{59.86}\\
    \bottomrule
    \end{tabular}
\end{table}

\subsection{3D Pose Estimation with HybrIK-Transformer}
HybrIK is a state-of-the-art single-view 3D HPE framework.
It uses a CNN and a hybrid inverse kinematics (HybrIK) algorithm of SMPL for joint 3D keypoints and human mesh retrieval \cite{li2021hybrik}.
HybrIK-Transformer replaces the deconvolution head in HybrIK with an attention module for 3D pose regression.
Compared to HybrIK, it achieves higher estimation accuracy and lower computational cost at the same time  \cite{oreshkin2023hybrikT}.
The HybrIK-Transformer (HybrIK-T.) model with the HRNet-48W backbone \cite{Sun2019hrnet} reaches a mean per-joint position error (MPJPE) of 71.6\,mm and a procrustes aligned MPJPE (PA-MPJPE) of 42.3\,mm on the 3DPW dataset \cite{vonMarcard20183dpw}, and an MPJPE of 47.5\,mm, a PA-MPJPE of 29.5\,mm on Human3.6M.
We use this model as the baseline and finetune it on NToP, PanopTOP31K, and THEODORE+.
As there is no real-world top-view 3D human pose dataset available, we evaluate the finetuned models on the validation sets of NToP and PanopTOP31K, presenting the results in \cref{tab:hybrikval}.

We make the following observations from the results:
1) When applied for top-view HPE, the error of the baseline model increases by over 100\,\% compared to the values of side-view. It is worth noting that the original training data includes MPI-INF-3DHP, so the model has seen overhead images.
2) Finetuning on \emph{any} of the train subsets of NToP improves performance on both evaluation sets.
3) Finetuning on PanopTOP31K does not improve results for NTOP-val, indicating low generalization ability. This achieves the best PA-MPJPE and MPJPE values on PanopTOP31K val, but PCK and AUC remain lower than those obtained by finetuning on NToP-train.
4) Finetuning on THEODORE+ shows worst performance, with no improvement for NToP-val and marginal improvement for PanopTOP31K val.
In summary, our dataset demonstrates superior generalization ability compared to existing 3D top-view HPE datasets.

\begin{table}[t]
    \centering
    \scriptsize
    \caption{Evaluation results of HybrIK-T. model. PA-MPJPE (\mbox{PA-M.}) and MPJPE are calculated for 24 joints of NToP-val and 15 joints of \mbox{PanopTOP31K} val. Their values are in \(\mathrm{mm}\), lower is better. We also calculate percentage of correct keypoints (PCK) and area under the curve (AUC), higher is better.}
    \label{tab:hybrikval}
    \begin{tabular}{lC{1cm}C{1cm}C{1cm}C{1cm}C{0.1cm}C{1cm}C{1cm}C{1cm}C{1cm}}
    \cmidrule[1pt]{2-10}
        & \multicolumn{4}{c}{NToP-val} & &\multicolumn{4}{c}{PanopTOP31K val} \\
    \cmidrule{2-10}
    Training set & PA-M.\textdownarrow & MPJPE\textdownarrow & PCK\textuparrow & AUC\textuparrow & & PA-M.\textdownarrow & MPJPE\textdownarrow & PCK\textuparrow & AUC\textuparrow \\
    \midrule
    Baseline HybrIK-T. & 115.5 & 136.0 & 55.03 & 36.42 & & 64.9 & 166.5 & 78.28 & 54.46 \\
    \midrule
    THEODORE+ & 122.9 & 115.3 & 62.31 & 30.57 & & 57.6 & 105.2 & 85.65 & 69.73 \\
    PanopTOP31K train & 113.5 & 156.2 & 64.64 & 32.91 & & \textbf{45.2} & \textbf{85.3} & 97.29 & 74.41 \\
    \midrule
    ntopH36M-train & 67.5 & 95.6 & 99.12 & 98.26 & & 52.3 & 98.6 & 97.64 & 73.77 \\
    ntopGB-train & 87.7 & 114.9 & 92.63 & 91.32 & & 54.2 & 99.3 & 92.18 & 68.15 \\
    ntopZJU-train & 87.7 & 114.9 & 92.91 & 91.57 & & 54.2 & 99.3 & 91.15 & 67.43 \\
    \textbf{NToP-train} & \textbf{61.8} & \textbf{88.9} & \textbf{99.67} & \textbf{98.72} & & 49.1 & 95.1 & \textbf{98.30} & \textbf{74.62}\\
    \bottomrule
    \end{tabular}
\end{table}

%% file: sections/discussion.tex
\section{Discussion}
\begin{figure}[h]
    \centering
    \subcaptionbox{\label{subfig:artifact}}[.22\linewidth]{\includegraphics[width=\linewidth]{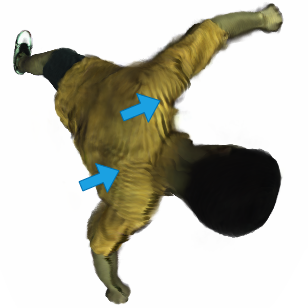}}\hspace{2mm}
    \subcaptionbox{\label{subfig:artifact2}}[.22\linewidth]{\includegraphics[width=\linewidth]{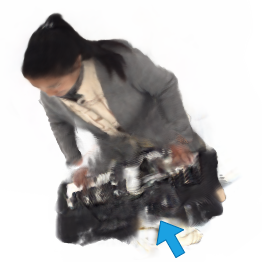}}\hspace{2mm}
    \subcaptionbox{\label{subfig:artifact3}}[.22\linewidth]{\includegraphics[width=\linewidth]{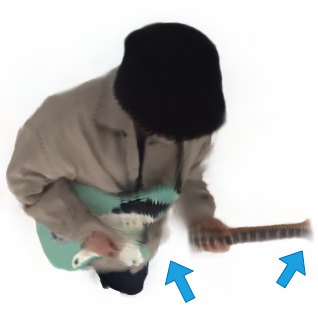}}\hspace{2mm}
    \subcaptionbox{\label{subfig:artifact4}}[.22\linewidth]{\includegraphics[width=\linewidth]{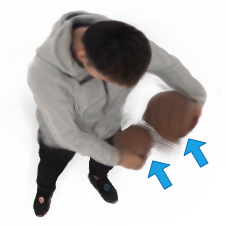}}
    \caption{(a) Stripe Artifacts due to non-linear ray distribution in omnidirectional rendering: actor p393 from zjumocap subset, frame 75, \(h=1.0\), \(R=0.5\), camera position NW. (b-d) Artifacts caused by incorrect modelling in genebody subset: (b) The keyboard is broken in the middle. (c) Parts of the guitar are missing. (d) The basketball is shown on both hands.}
    \label{fig:artifact1}
\end{figure}

\textbf{Render artifacts.}
There are two main reasons that result in artifacts in the rendered images.
The first source of artifacts is the inherent non-linearity of the ray finding process for the virtual fisheye camera, see \cref{subfig:rays}.
Artifacts of such source show as stripes in the rendered images, see \cref{subfig:artifact}.
The artifacts become more prominent as the render size increases.
This problem can be partially reduced by training with higher resolution images and keeping the rendered subjects smaller than in the origin dataset.
Due to this problem, the human36m subset is not rendered with \(h=1.0\) and \(R=0.5\), as the subjects are larger than in the training images in this setting, and the artifacts become unacceptable.

The second source of artifacts is non-human objects in the scene.
Since SMPL only explicitly model the human body, objects that are very close to the actor are implicitly modeled as deformable part of the body, but objects that are farther away cannot be modeled correctly.
\Cref{fig:artifact1}\red{b-d} shows a few examples.
A related issue is that the model cannot be trained properly if major parts of the body are covered by objects or large pieces of clothes.
For example, if the actor wears a long, puffy dress, and the legs and feet are not visible, training cannot converge due to ambiguity in the lower body SMPL mesh.


\textbf{Computational cost.}
In the current implementation of the NToP pipeline, one model must be trained for each actor, and training one model takes 24 hours in average with our setup of two Nvidia TITAN RTX GPUs.
Rendering with the same hardware takes 4\,\(\sim\)\,7 seconds per image, depending on the rendered subject size.
Therefore, the total computation time for NToP amounts to 80 days.

\textbf{Future perspectives.}
We have solely utilized NToP for single-view 2D and 3D pose estimation in this study, leaving numerous possibilities for future exploration.
The multiple viewpoints in our rendering enable the utilization of multi-view pose estimation techniques in top-view scenarios.
Retaining temporal information from the origin datasets in NToP could enhance the precision of pose estimation and facilitate action recognition in top-view settings.
Moreover, the virtual camera setup can be readily adapted to generate datasets for a stereo camera rig or for creating egocentric pose estimation datasets \cite{yu2023survey}.
On another front, the NToP stands to benefit from enhancements by incorporating features from novel human-centric NeRF models \cite{Chen2023CVPR,wang2023clothednerf} to reduce the computational costs and leveraging emerging high-resolution motion capture datasets \cite{cheng2023dnarendering} to further enhance the render quality.
Our next step involves adopting Gaussian splatting scene rendering \cite{kerbl3Dgaussians} into the NToP pipeline, which will enable us to embed scenes and further enhance the realism and diversity of the dataset.

%% file: sections/conclusion.tex
\section{Conclusion}
In this paper, we present the dataset generation pipeline that is capable of rendering large-scale top-view omnidirectional HPE datasets with groundtruth keypoint annotations.
Furthermore, our pipeline provides camera parameters and segmentation masks.
Using the pipeline, we generate a novel dataset NToP (NeRF-powered Top-view human Pose dataset for fisheye cameras), and prove its effectiveness for HPE in top-view images by finetuning and testing the 2D pose estimation model ViTPose-B and the 3D pose estimator HybrIK-Transformer.
Our results show that NToP is effective for 2D and 3D top-view HPE and prove its superiority over existing datasets in this field.

Our new pipeline enables researchers to make use of existing side-view HPE datasets to overcome the difficulty of acquiring high-quality top-view data for HPE. This opens up numerous possibilities for the application of top-view fisheye cameras, for instance human action and behaviour analysis and accurate emergency detection.

%% file: sections/supp.tex
\renewcommand\thesection{\Alph{section}} 
\setcounter{section}{0}
\section{Supplementary Material}

\subsection{Extended Evaluation Results}
We provide extended evaluation results on the validation subsets of NToP of the finetuned models.
In addition to the results presented in Tab. 3, the baseline and finetuned ViTPose-B models are evaluated on the subsets of NToP-val as well, see \cref{tab:vitvalsupp}.
Aside from the observations in the paper,  we notice that
finetuning on ntopGB-train results in the smallest performance drop on OmniLab and PoseFES, which means it has the smallest domain gap to real-world images.
The most probable reason is the diversity of actors and the high recording quality.
Conversely, the model finetuned with NToP-train and COCO performs the best on ntopGB-val, while for ntopH36M-val and notpZJU-val, the best performing models are those without COCO.
This further proves the quality of notpGB.
Based on this, we perform a further experiment and finetune ViTPose-B again on ntopGB and COCO together. 
However, the finetuning converges at AP=78.11\% and AR=80.18\% for OmniLab, which is still lower than NToP-train and COCO.
This suggests that the diversity in NToP-train is helpful for model generation despite of the domain gap.

\begin{table}[h]
    \centering
    \scriptsize
    \caption{Full evaluation results of 2D keypoints with ViTPose-B model on NToP-val and OmniLab. AP and AR are calculated following the MS COCO evaluation pipeline, presented in \%. T+ stands for THEODORE+.}
    \label{tab:vitvalsupp}
    \begin{tabular}{lcccccccccccc}
    \cmidrule[1pt]{2-13}
     & \multicolumn{2}{c}{ntopH36M-val} & \multicolumn{2}{c}{ntopGB-val} & \multicolumn{2}{c}{ntopZJU-val} & \multicolumn{2}{c}{NToP-val} & \multicolumn{2}{c}{OmniLab} & \multicolumn{2}{c}{PoseFES} \\ 
     \cmidrule{2-13}
    Training set & AP    & AR   & AP    & AR   & AP    & AR   & AP   & AR     & AP   & AR  & AP   & AR\\
    \midrule
    baseline ViTP-B & 50.12 & 53.37 & 23.45 & 30.01 & 43.96 & 47.60 & 46.36 & 49.82 & 76.62 & 78.53 & 54.20 & 57.70\\
    \midrule
    T+ & 59.50 & 62.83 & 22.30 & 28.10 & 57.96 & 61.21 & 54.44 & 57.74 & 69.14 & 72.56 & 46.90 & 51.23\\
    T+\&coco & 65.16 & 68.27 & 28.79 & 35.64 & 65.64 & 69.12 & 60.27 & 63.55 & 76.52 & 79.02 & 55.58 & 58.82\\    
    \midrule
    ntopH36M-train & 81.94 & 84.26 & 23.80 & 30.79 & 61.26 & 63.98 & 73.51 & 75.93 & 61.32 & 64.95 & 36.42 & 40.54\\
    ntopGB-train & 66.59 & 70.11 & 46.97 & 54.63 & 64.33 & 67.94 & 63.60 & 67.60 & 74.95 & 77.71 & 53.56 & 57.44 \\
    ntopZJU-train & 60.31 & 64.51 & 40.12 & 48.24 & \textbf{85.61} & \textbf{87.89} & 57.15 & 61.87 & 69.58 & 73.59 & 52.38 & 46.48\\
    NToP-train & \textbf{84.55} & \textbf{86.72} & 47.89 & 55.37 & 83.00 & 85.85 & \textbf{79.71} & \textbf{82.07} & 70.19 & 73.49 & 46.84 & 51.34\\
    NToP-train\&coco  & 83.86 & 86.28 & \textbf{49.14} & \textbf{56.75} & 81.99 & 84.71 & 79.54 & 81.95 & \textbf{78.25} & \textbf{80.51} & \textbf{56.47} & \textbf{59.86}\\
    \bottomrule
    \end{tabular}
\end{table}

We evaluate HybrIK-Transformer with a similar scheme and present the results in \cref{tab:hybrikvalsupp,tab:hybrik_metrics}.
Due to the lack of real-world data, we are unable to make similar observations like for ViTPose.
In the 3D case, finetuning on NToP-train generally delivers the best performance, except PA-MPJPE and MPJPE for PT31K, which is expected.
THEODORE+, on the other hand, hardly improves model performance on either NToP-val or PanopTOP31K.
What is interesting and noteworthy is the fact that neither adding THEODORE+ nor adding PanopTOP31K to NToP-train improves the performance for NToP-val or PanopTOP31K val.
We also train HybrIK-Transformer from scratch with NToP-train.
The trained model performs better than the baseline on both evaluation sets, yet it can not compete with finetuning the baseline model on NToP-train.

\begin{table}[h]
    \centering
    \scriptsize
    \caption{Evaluation results of HybrIK-Transformer model on NToP and \mbox{PanopTOP31K} (PT31K). PA-MPJPE (PA-M.) and MPJPE are calculated for 24 SMPL joints for NToP and 15 joints for PanopTOP31K. Values are in \(\mathrm{mm}\), lower is better. NToP-train(scratch) denotes results of training from scratch on NToP-train.}
    \label{tab:hybrikvalsupp}
    \begin{tabular}{lcccccccccc}
    \cmidrule[1pt]{2-11}
     & \multicolumn{2}{c}{ntopH36M-val} & \multicolumn{2}{c}{ntopGB-val} & \multicolumn{2}{c}{ntopZJU-val} & \multicolumn{2}{c}{NToP-val} & \multicolumn{2}{c}{PT31K val} \\ 
     \cmidrule{2-11} 
     & PA-M. & MPJPE & PA-M. & MPJPE & PA-M. & MPJPE & PA-M. & MPJPE & PA-M. & MPJPE \\
    \midrule
    baseline HybrIK-T & 119.1 & 138.7 & 98.2 & 122.6 & 98.7 & 156.1 & 115.5 & 136.0 & 64.9 & 166.5\\
    \midrule
    THEODORE+ & 117.5 & 156.2 & 98.1 & 154.2 & 97.2 & 151.2 & 122.9 & 115.3 & 57.6 & 105.2 \\
    PT31K & 105.1 & 112.17 & 89.9 & 130.1 & 91.4 & 128.6 & 113.5 & 156.2 & \textbf{45.2} & \textbf{85.3}\\
    \midrule
    ntopH36M-train & 67.0 & 91.9 & 61.7 & 100.3 & 69.4 & 113.5 & 67.5 & 95.6 & 52.3 & 98.6\\
    ntopGB-train & 82.3 & 113.7 & 69.2 & 108.7 & 78.8 & 123.8 & 87.7 & 114.9 & 54.2 & 99.3 \\
    ntopZJU-train & 82.0 & 114.1 & 73.8 & 115.9 & 72.6 & 116.3 & 81.9 & 114.0 & 52.8 & 98.7 \\
    NToP-train & \textbf{62.7} & \textbf{87.1} & \textbf{58.3} & \textbf{97.2} & \textbf{64.2} & \textbf{105.5} & \textbf{61.8} & \textbf{88.9} & 49.1 & 95.1\\
    NToP-train\&T+ & 75.7 & 105.9 & 68.8 & 101.8 & 77.3 & 111.3 & 74.4 & 103.3 & 55.9 & 100.4\\
    NToP-train\&PT31K & 74.3 & 98.4 & 68.5 & 98.2 & 73.4 & 103.7 & 72.1 & 98.2 & 53.1 & 98.1\\
    NToP-train(scratch) & 92.7 & 117.1 & 77.1 & 124.2 & 85.9 & 126.9 & 91.5 & 115.2 & 57.3 & 102.0\\
    
    \bottomrule
    \end{tabular}
\end{table}

\begin{table}[h]
    \centering
    \scriptsize
    \caption{PCK and AUC values. NToP-train(scratch) denotes results of training from scratch on NToP-train.}
    \label{tab:hybrik_metrics}
    \begin{tabular}{lC{1cm}C{1cm}C{1cm}C{1cm}C{1cm}C{1cm}C{1cm}C{1cm}}
     \cmidrule[1pt]{2-5}
     & \multicolumn{2}{c}{NToP-val (24 Jts)} & \multicolumn{2}{c}{PT31K val (15 Jts)} \\ 
     \cmidrule{2-5} 
        & PCK & AUC & PCK & AUC \\
     \midrule
    baseline HybrIK & 55.03 & 36.42 & 78.28 & 54.46 \\
    \midrule
    Theodore & 62.31 & 30.57 & 85.65 & 69.73\\
    PT31K & 64.64 & 32.91 & 97.29 & 74.41\\
    \midrule
    NToPH36M-train & 99.12 & 98.26 & 97.64 & 73.77\\
    NToPGB-train & 92.63 & 91.32 & 92.18 & 68.15\\
    NToPZJU-train & 92.91 & 91.57 & 91.15 & 67.43\\
    NToP-train & \textbf{99.67} & \textbf{98.72} & \textbf{98.30} & \textbf{74.62}\\
    NToP \& Theodore & 96.34 & 95.73 & 92.67 & 70.28\\
    NToP \& PT31K & 96.89 & 95.77 & 95.79 & 70.54\\
    NToP-train(scratch) & 90.92 & 89.15 & 87.69 & 65.33\\
    \bottomrule
    \end{tabular}
\end{table}

\subsection{NToP Samples and Render Setting Comparison}
\begin{figure}
    \centering
    \subcaptionbox{\label{subfig:r10050}}[.23\linewidth]{\includegraphics[width=\linewidth]{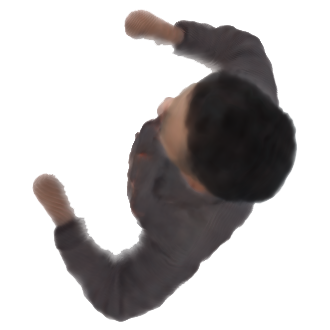}}\hspace{2mm}
    \subcaptionbox{\label{subfig:r10051}}[.23\linewidth]{\includegraphics[width=\linewidth]{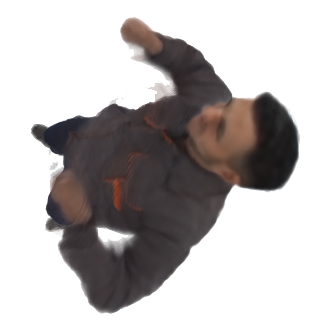}}\hspace{2mm}
    \subcaptionbox{\label{subfig:r10056}}[.23\linewidth]{\includegraphics[width=\linewidth]{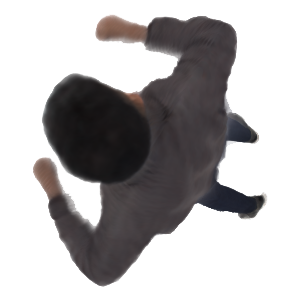}}\hspace{2mm}
    \subcaptionbox{\label{subfig:r10057}}[.23\linewidth]{\includegraphics[width=\linewidth]{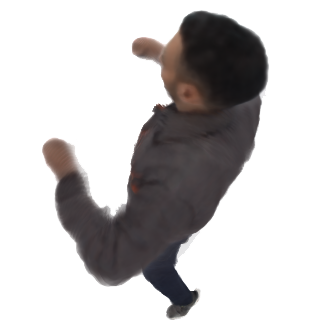}}
    \subcaptionbox{\label{subfig:r12100}}[.23\linewidth]{\includegraphics[width=\linewidth]{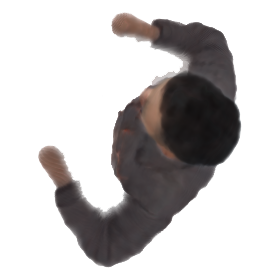}}\hspace{2mm}
    \subcaptionbox{\label{subfig:r12101}}[.23\linewidth]{\includegraphics[width=\linewidth]{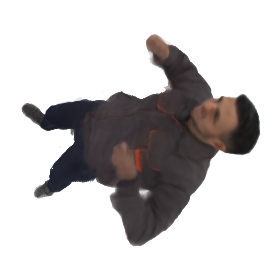}}\hspace{2mm}
    \subcaptionbox{\label{subfig:r12106}}[.23\linewidth]{\includegraphics[width=\linewidth]{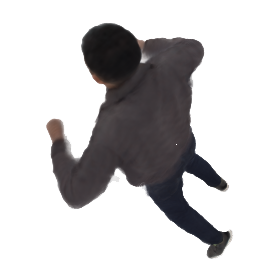}}\hspace{2mm}
    \subcaptionbox{\label{subfig:r12107}}[.23\linewidth]{\includegraphics[width=\linewidth]{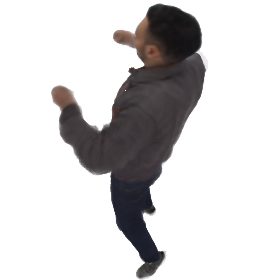}}
    \caption{Rivera in ntopGB. Top row (a-d) shows camera at C, W, SE and S with H=1.0 and R=0.5, bottom row (e-h) shows camera at same positions with H=1.2 and R=1.0.}
    \label{fig:hrrivera}
\end{figure}

\begin{figure}
    \centering
    \subcaptionbox{\label{subfig:x10052}}[.23\linewidth]{\includegraphics[width=\linewidth]{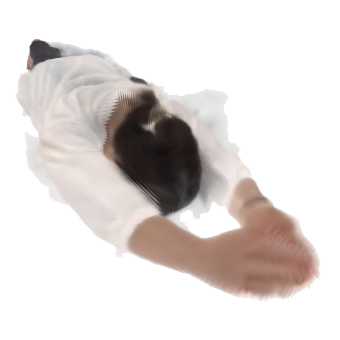}}\hspace{2mm}
    \subcaptionbox{\label{subfig:x10053}}[.23\linewidth]{\includegraphics[width=\linewidth]{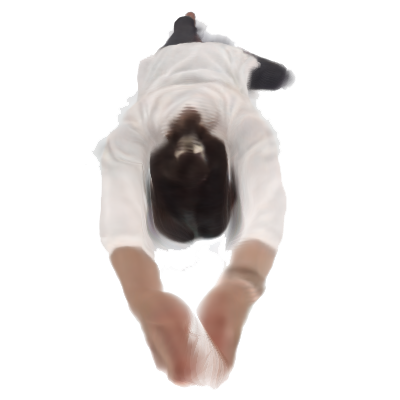}}\hspace{2mm}
    \subcaptionbox{\label{subfig:x10054}}[.23\linewidth]{\includegraphics[width=\linewidth]{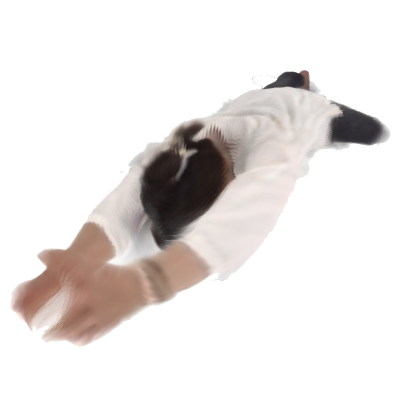}}\hspace{2mm}
    \subcaptionbox{\label{subfig:x10055}}[.23\linewidth]{\includegraphics[width=\linewidth]{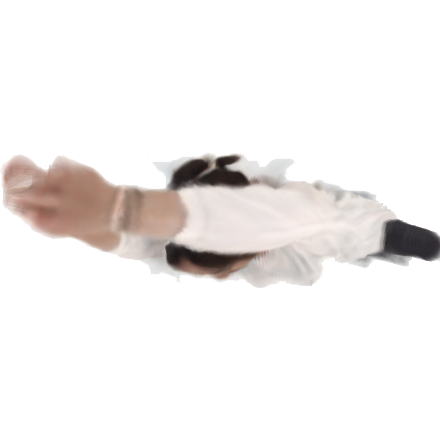}}
    \subcaptionbox{\label{subfig:x12102}}[.23\linewidth]{\includegraphics[width=\linewidth]{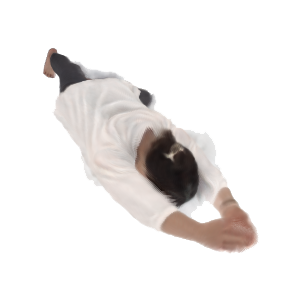}}\hspace{2mm}
    \subcaptionbox{\label{subfig:x12103}}[.23\linewidth]{\includegraphics[width=\linewidth]{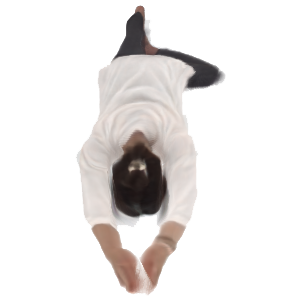}}\hspace{2mm}
    \subcaptionbox{\label{subfig:x12104}}[.23\linewidth]{\includegraphics[width=\linewidth]{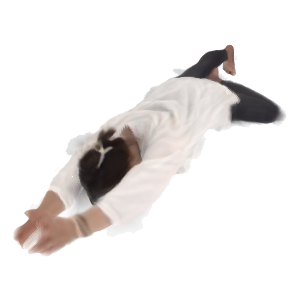}}\hspace{2mm}
    \subcaptionbox{\label{subfig:x12105}}[.23\linewidth]{\includegraphics[width=\linewidth]{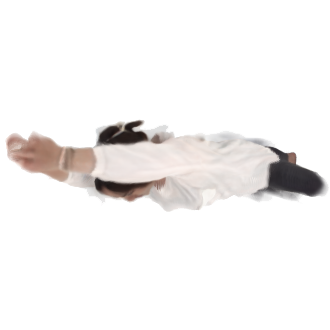}}
    \caption{Xujiarui in ntopGB. Top row (a-d) shows camera at NW, N, NE and E with H=1.0 and R=0.5, bottom row (e-h) shows camera at same positions with H=1.2 and R=1.0.}
    \label{fig:hrxujiarui}
\end{figure}

The top-view image of fisheye cameras changes drastically as distance of the subject of interest to the camera optical axis changes.
Subjects, that are far away from the camera, are very similar to normal perspective, therefore trivial for person detectors and keypoints estimators due to their generalization ability, despite the difference in orientation.
Yet those close to the optical axis changes their appearance drastically, and the lateral distance as well as the height of the camera has a huge impact on the occlusion and appearance.
Therefore, we render with different H and R for the virtual camera.
\cref{fig:hrrivera,fig:hrxujiarui} showcase the effect of H and R for the rendering.
Furthermore, we provide a supplementary video to showcase the variability of NToP dataset.

\begin{figure}
    \centering
    \includegraphics[width=.19\linewidth]{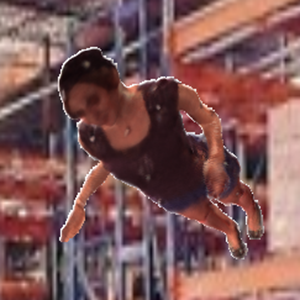}\vspace{2pt}
    \includegraphics[width=.19\linewidth]{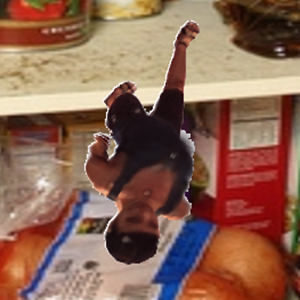}
    \includegraphics[width=.19\linewidth]{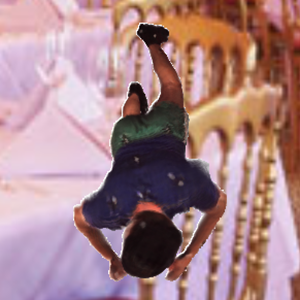}
    \includegraphics[width=.19\linewidth]{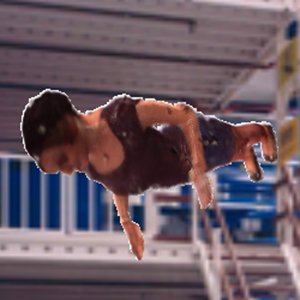}
    \includegraphics[width=.19\linewidth]{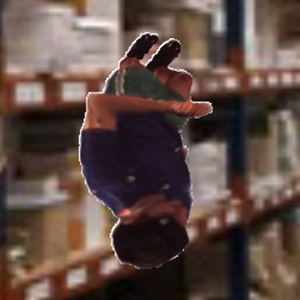}
    \includegraphics[width=.19\linewidth]{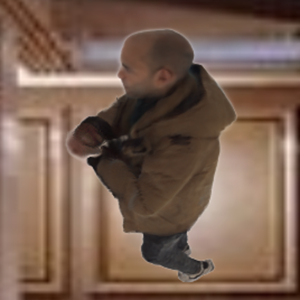}\vspace{2pt}
    \includegraphics[width=.19\linewidth]{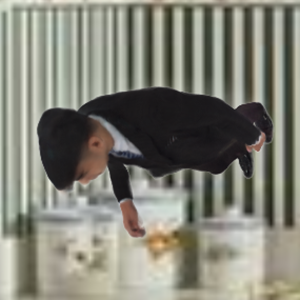}
    \includegraphics[width=.19\linewidth]{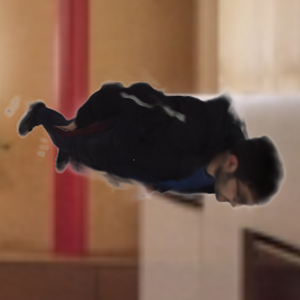}
    \includegraphics[width=.19\linewidth]{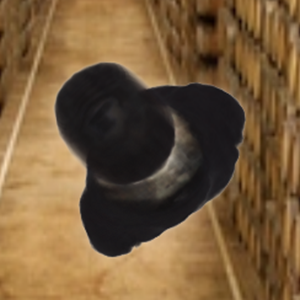}
    \includegraphics[width=.19\linewidth]{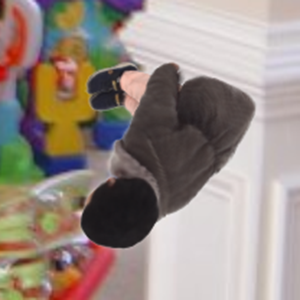}
    \includegraphics[width=.19\linewidth]{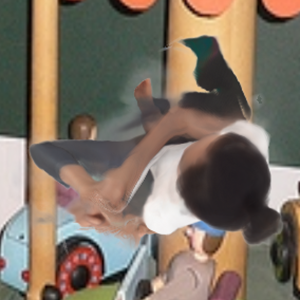}
    \includegraphics[width=.19\linewidth]{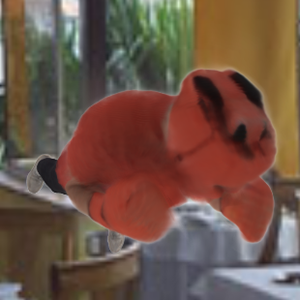}
    \includegraphics[width=.19\linewidth]{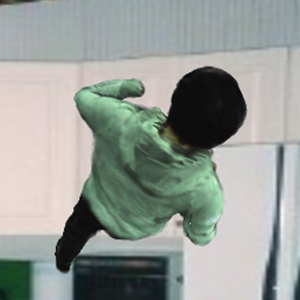}
    \includegraphics[width=.19\linewidth]{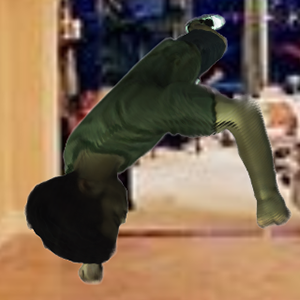}
    \includegraphics[width=.19\linewidth]{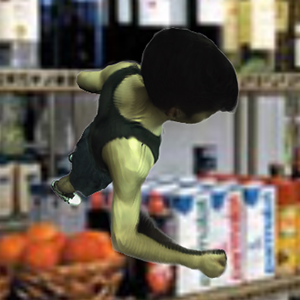}
    \caption{Random examples of NToP with background image from MIT Indoor Scenes database.}
    \label{fig:ntopwbg}
\end{figure}

To further increase the realness and variety of NToP and decrease the domain gap between NToP and real-world data, we augment the background of NToP using MIT Indoor Scenes database.
The original MIT Indoor Scenes images are reduced to 9400 images without human subjects, then randomly chosen as backgrounds for NToP.
Some random examples after background augmentation are presented in \cref{fig:ntopwbg}.

\subsection{OmniLab}
\begin{figure}
    \centering
    \subcaptionbox{Livingroom\label{subfig:livingroom}}[.49\linewidth]{\includegraphics[width=\linewidth]{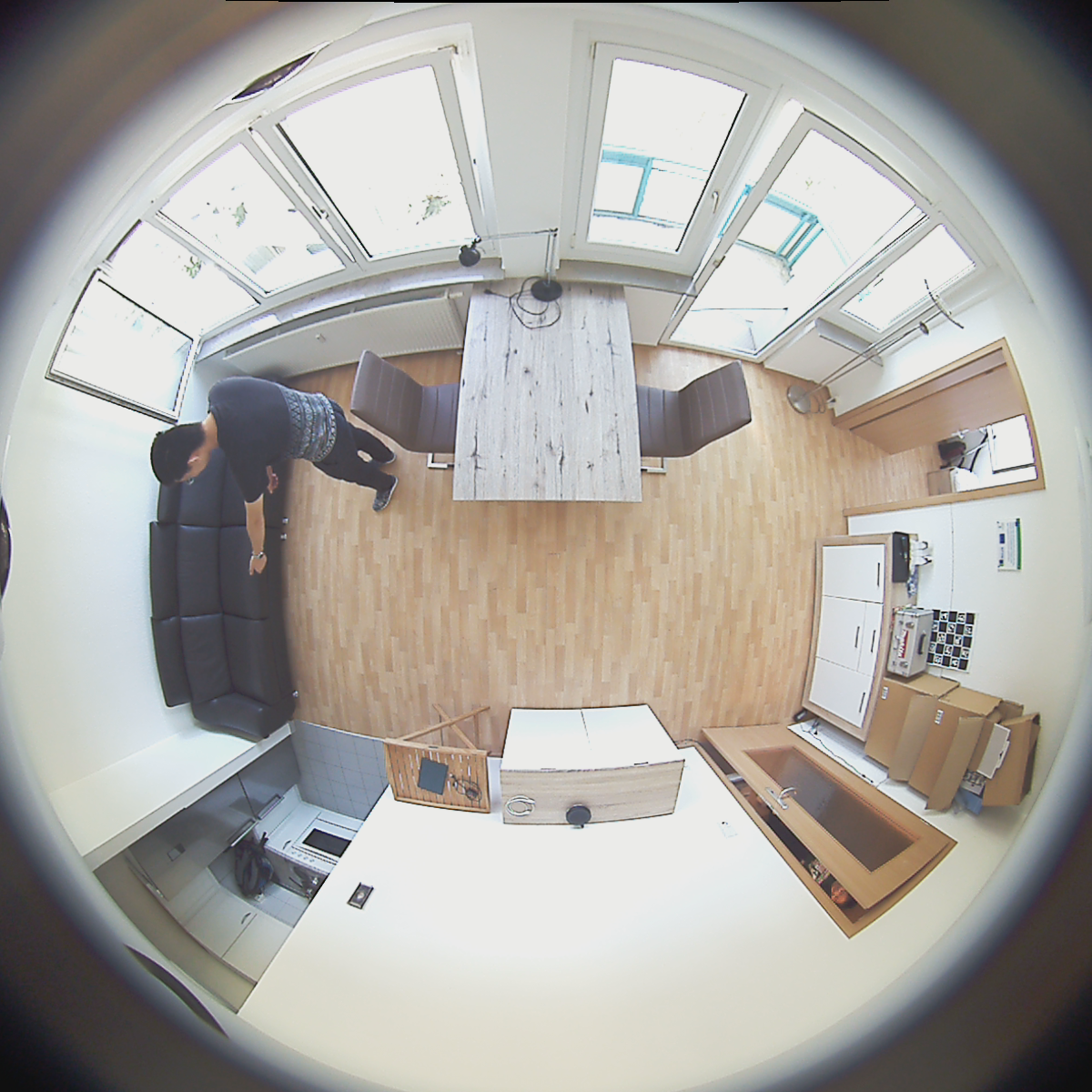}}
    \subcaptionbox{bedroom\label{subfig:bedroom}}[.49\linewidth]{\includegraphics[width=\linewidth]{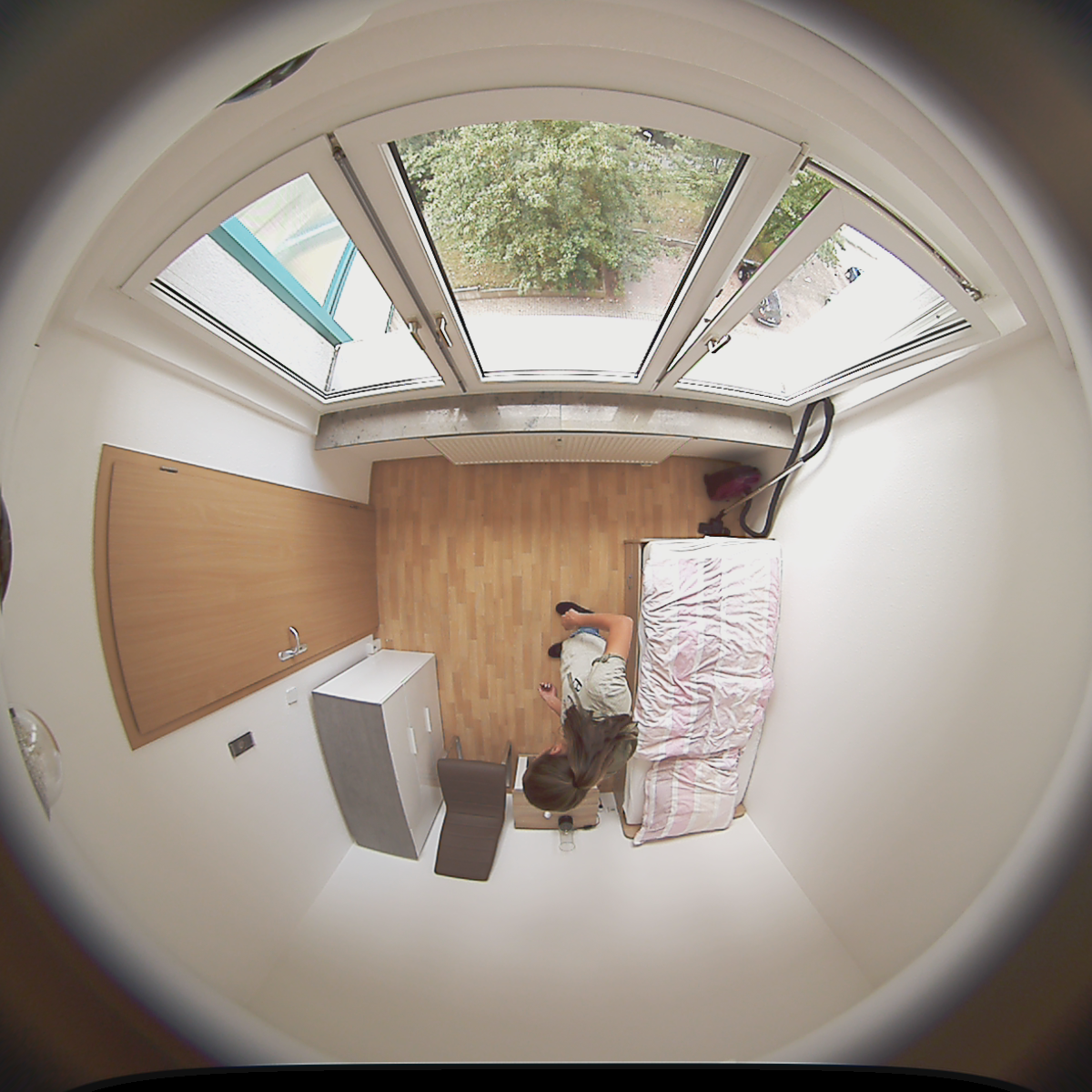}}
    \caption{Room used for recording OmniLab.}
    \label{fig:omniroom}
\end{figure}

\begin{figure}
    \centering
    \subcaptionbox{brooming\label{subfig:brooming}}[.235\linewidth]{\includegraphics[width=\linewidth]{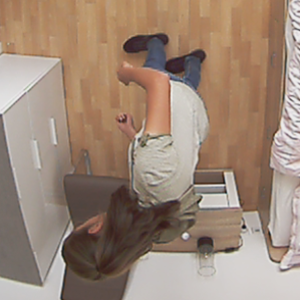}}\hspace{1mm}
    \subcaptionbox{cleaning windows\label{subfig:cleaningw}}[.235\linewidth]{\includegraphics[width=\linewidth]{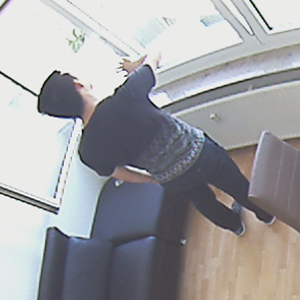}}\hspace{1mm}
    \subcaptionbox{down and get up\label{subfig:downgetup}}[.235\linewidth]{\includegraphics[width=\linewidth]{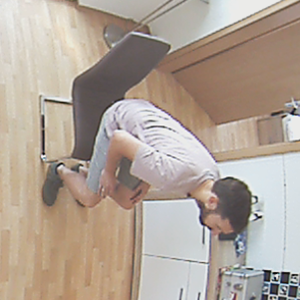}}\hspace{1mm}
    \subcaptionbox{drinking\label{subfig:drinking}}[.235\linewidth]{\includegraphics[width=\linewidth]{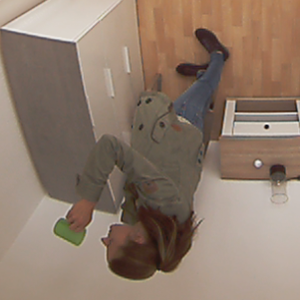}}
    
    \subcaptionbox{fall on face\label{subfig:fallonface}}[.235\linewidth]{\includegraphics[width=\linewidth]{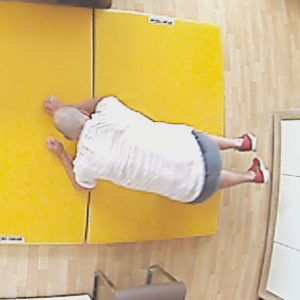}}\hspace{1mm}
    \subcaptionbox{in chair stand up\label{subfig:inchairstand}}[.235\linewidth]{\includegraphics[width=\linewidth]{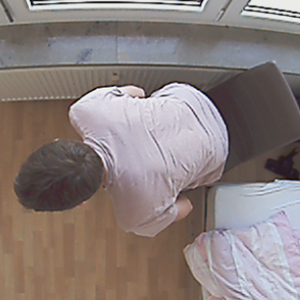}}\hspace{1mm}
    \subcaptionbox{pull object\label{subfig:pullobject}}[.235\linewidth]{\includegraphics[width=\linewidth]{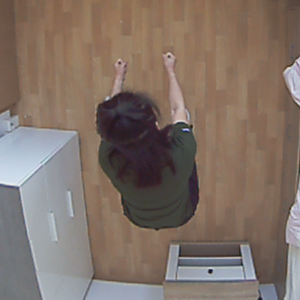}}\hspace{1mm}
    \subcaptionbox{push object\label{subfig:pushobject}}[.235\linewidth]{\includegraphics[width=\linewidth]{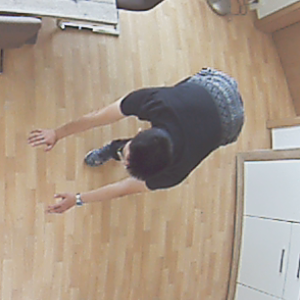}}
    
    \subcaptionbox{rugpull\label{subfig:rugpull}}[.235\linewidth]{\includegraphics[width=\linewidth]{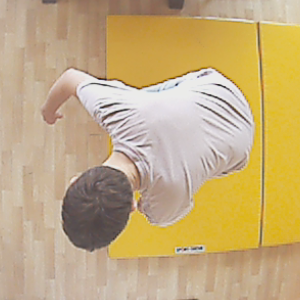}}\hspace{1mm}
    \subcaptionbox{turn left\label{subfig:turnleft}}[.235\linewidth]{\includegraphics[width=\linewidth]{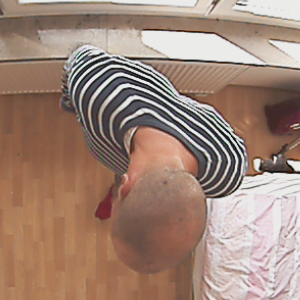}}\hspace{1mm}
    \subcaptionbox{turn right\label{subfig:turnright}}[.235\linewidth]{\includegraphics[width=\linewidth]{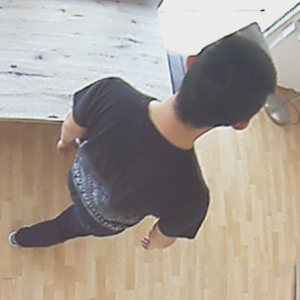}}\hspace{1mm}
    \subcaptionbox{upbend from knees\label{subfig:upfrmknees}}[.235\linewidth]{\includegraphics[width=\linewidth]{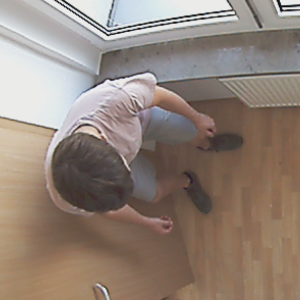}}
    
    \subcaptionbox{upbend from waist\label{subfig:upfrmwaist}}[.235\linewidth]{\includegraphics[width=\linewidth]{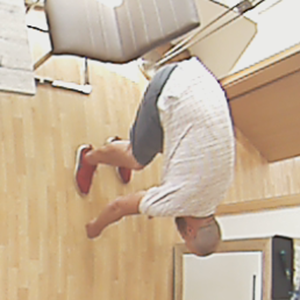}}\hspace{1mm}
    \subcaptionbox{up from ground\label{subfig:upfrmground}}[.235\linewidth]{\includegraphics[width=\linewidth]{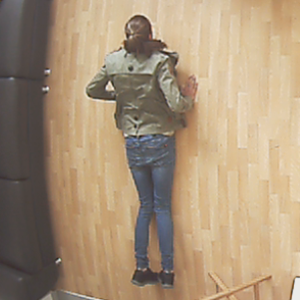}}\hspace{1mm}
    \subcaptionbox{walk\label{subfig:walk}}[.235\linewidth]{\includegraphics[width=\linewidth]{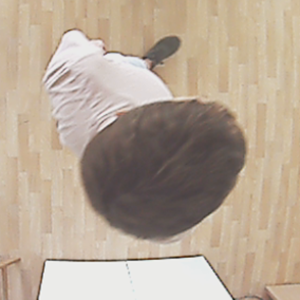}}\hspace{1mm}
    \subcaptionbox{walk-old-man\label{subfig:walkoldman}}[.235\linewidth]{\includegraphics[width=\linewidth]{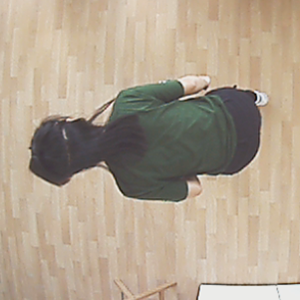}}
    \caption{Examples for each action in OmniLab}
    \label{fig:omnilabaction}
\end{figure}

OmniLab dataset is recorded in our apartment lab, which is equipped with one fisheye camera in each room.
We used the living room and the bedroom for recording, see \cref{fig:omniroom}.
The actions performed in the recordings are: brooming, cleaning windows, down and get up, drinking, fall-on-face, in chair and stand up, pull object, push object, rugpull, turn left, turn right, upbend from knees, upbend from waist, up from ground, walk, walk-old-man.
We present one example from each action in \cref{fig:omnilabaction}, the examples are deliberately chosen to not reveal the faces to preserve anonymity. 

